\PassOptionsToPackage{unicode}{hyperref}
\PassOptionsToPackage{hyphens}{url}
\PassOptionsToPackage{dvipsnames,svgnames,x11names}{xcolor}
\documentclass[12pt]{article}

\usepackage{amsmath,amssymb}
\usepackage{iftex}
\ifPDFTeX
  \usepackage[T1]{fontenc}
  \usepackage[utf8]{inputenc}
  \usepackage{textcomp} 
\else 
  \usepackage{unicode-math}
  \defaultfontfeatures{Scale=MatchLowercase}
  \defaultfontfeatures[\rmfamily]{Ligatures=TeX,Scale=1}
\fi
\usepackage{lmodern}
\ifPDFTeX\else  
\fi
\IfFileExists{upquote.sty}{\usepackage{upquote}}{}
\IfFileExists{microtype.sty}{
  \usepackage[]{microtype}
  \UseMicrotypeSet[protrusion]{basicmath} 
}{}
\makeatletter
\@ifundefined{KOMAClassName}{
  \IfFileExists{parskip.sty}{%
    \usepackage{parskip}
  }{
    \setlength{\parindent}{0pt}
    \setlength{\parskip}{6pt plus 2pt minus 1pt}}
}{
  \KOMAoptions{parskip=half}}
\makeatother
\usepackage{xcolor}
\makeatletter
\ifx\paragraph\undefined\else
  \let\oldparagraph\paragraph
  \renewcommand{\paragraph}{
    \@ifstar
      \xxxParagraphStar
      \xxxParagraphNoStar
  }
  \newcommand{\xxxParagraphStar}[1]{\oldparagraph*{#1}\mbox{}}
  \newcommand{\xxxParagraphNoStar}[1]{\oldparagraph{#1}\mbox{}}
\fi
\ifx\subparagraph\undefined\else
  \let\oldsubparagraph\subparagraph
  \renewcommand{\subparagraph}{
    \@ifstar
      \xxxSubParagraphStar
      \xxxSubParagraphNoStar
  }
  \newcommand{\xxxSubParagraphStar}[1]{\oldsubparagraph*{#1}\mbox{}}
  \newcommand{\xxxSubParagraphNoStar}[1]{\oldsubparagraph{#1}\mbox{}}
\fi
\makeatother

\usepackage{longtable,booktabs,array}
\usepackage{calc} 
\usepackage{etoolbox}
\makeatletter
\patchcmd\longtable{\par}{\if@noskipsec\mbox{}\fi\par}{}{}
\makeatother
\IfFileExists{footnotehyper.sty}{\usepackage{footnotehyper}}{\usepackage{footnote}}
\makesavenoteenv{longtable}
\usepackage{graphicx}
\makeatletter
\def\maxwidth{\ifdim\Gin@nat@width>\linewidth\linewidth\else\Gin@nat@width\fi}
\def\maxheight{\ifdim\Gin@nat@height>\textheight\textheight\else\Gin@nat@height\fi}
\makeatother
\setkeys{Gin}{width=\maxwidth,height=\maxheight,keepaspectratio}
\makeatletter
\def\fps@figure{htbp}
\makeatother

\makeatletter
\@ifpackageloaded{caption}{}{\usepackage{caption}}
\AtBeginDocument{%
\ifdefined\contentsname
  \renewcommand*\contentsname{Table of contents}
\else
  \newcommand\contentsname{Table of contents}
\fi
\ifdefined\listfigurename
  \renewcommand*\listfigurename{List of Figures}
\else
  \newcommand\listfigurename{List of Figures}
\fi
\ifdefined\listtablename
  \renewcommand*\listtablename{List of Tables}
\else
  \newcommand\listtablename{List of Tables}
\fi
\ifdefined\figurename
  \renewcommand*\figurename{Figure}
\else
  \newcommand\figurename{Figure}
\fi
\ifdefined\tablename
  \renewcommand*\tablename{Table}
\else
  \newcommand\tablename{Table}
\fi
}
\@ifpackageloaded{float}{}{\usepackage{float}}
\floatstyle{ruled}
\@ifundefined{c@chapter}{\newfloat{codelisting}{h}{lop}}{\newfloat{codelisting}{h}{lop}[chapter]}
\floatname{codelisting}{Listing}

\makeatother
\makeatletter
\@ifpackageloaded{caption}{}{\usepackage{caption}}
\@ifpackageloaded{subcaption}{}{\usepackage{subcaption}}
\makeatother

\ifLuaTeX
  \usepackage{selnolig}  
\fi
\usepackage[]{natbib}
\usepackage{bookmark}

\IfFileExists{xurl.sty}{\usepackage{xurl}}{} 
\hypersetup{
  pdftitle={StatEval: A Comprehensive Benchmark for Large Language Models in Statistics},
  pdfauthor={Yuchen Lu, Run Yang, Yichen Zhang, Shuguang Yu, Ziwei Wang, Jiayi Xiang, Wenxin E, Changyu Zhu, Fan Zhou},
  pdfkeywords={Statistical reasoning, Automated extraction, Proof verification, Multi-agent evaluation, Retrieval-augmented generation},
  colorlinks=true,
  linkcolor={blue},
  filecolor={Maroon},
  citecolor={Blue},
  urlcolor={Blue},
  pdfcreator={LaTeX via pandoc}}

\usepackage{psfrag,epsf}
\usepackage{enumerate}
\usepackage{multicol}
\usepackage{underscore}
\usepackage{listings} 
\usepackage{multirow}  
\usepackage{makecell}
\usepackage{tabularx}
\usepackage{enumitem} 
\usepackage{threeparttable}
\usepackage[framemethod=TikZ]{mdframed}
\renewcommand{\arraystretch}{1.2}
\mdfdefinestyle{journalprompt}{
    everyline=true,               
    linewidth=0.8pt,              
    linecolor=black!80,           
    backgroundcolor=black!2,      
    frametitlebackgroundcolor=black!65, 
    frametitlefont=\color{white}\bfseries\sffamily\footnotesize,
    innertopmargin=10pt,          
    innerbottommargin=10pt,
    innerleftmargin=8pt,
    innerrightmargin=8pt,
    splittopskip=1.2em,           
    splitbottomskip=1em,          
    font=\footnotesize,
    settings={\setlength{\parskip}{0pt}}, 
    needspace=1.5cm
}

\mdfdefinestyle{casestudyprompt}{
    everyline=true,               
    linewidth=0.8pt,              
    linecolor=Maroon!80,               
    backgroundcolor=Maroon!2,          
    frametitlebackgroundcolor=Maroon!75, 
    frametitlefont=\color{white}\bfseries\sffamily\footnotesize,
    innertopmargin=10pt,          
    innerbottommargin=10pt,
    innerleftmargin=8pt,
    innerrightmargin=8pt,
    splittopskip=1.2em,           
    splitbottomskip=1em,          
    font=\footnotesize,
    settings={\setlength{\parskip}{0pt}}, 
    needspace=1.5cm
}

\newcounter{promptnum}
\newcommand{\anon}{1}

\begin{document}

\def\spacingset#1{\renewcommand{\baselinestretch}%
{#1}\small\normalsize} \spacingset{1}


\if1\anon
{
  \title{\bf StatEval: A Comprehensive Benchmark for Large Language Models in Statistics}
  \author{
    Yuchen Lu\textsuperscript{*,1}, 
    Run Yang\textsuperscript{*,1}, 
    Yichen Zhang\textsuperscript{*,1}, 
    Shuguang Yu\textsuperscript{*,1}, \\
    Ziwei Wang\textsuperscript{1}, 
    Jiayi Xiang\textsuperscript{1}, 
    Wenxin E\textsuperscript{1}, 
    Changyu Zhu\textsuperscript{1}, 
    Fan Zhou\textsuperscript{\textdagger,1} \\
    \textsuperscript{1}Shanghai University of Finance and Economics
  }
  
  \date{} 
  
  \maketitle
  
  \begingroup
  \renewcommand{\thefootnote}{}%
  \footnotetext{\textsuperscript{*}Equal contribution}
  \footnotetext{\textsuperscript{\textdagger}Corresponding author: \texttt{zhoufan@mail.shufe.edu.cn}}
  \endgroup
} \fi

\if0\anon
{
  \bigskip
  \bigskip
  \bigskip
  \begin{center}
    {\LARGE\bf StatEval: A Comprehensive Benchmark for Large Language Models in Statistics}
\end{center}
  \medskip
} \fi

\bigskip
\begin{abstract}
Despite rapid advances in large language models (LLMs), statistical reasoning remains underrepresented in existing LLM benchmarks, which often do not reflect the layered, proof-driven nature of real statistical practice. To address this gap, we introduce \textbf{StatEval}, the first large-scale benchmark for statistical reasoning across curricular and research-level settings. StatEval includes over 100,000 curated problems, with 20,000+ foundational questions spanning undergraduate and graduate curricula and 80,000+ research-level proof tasks extracted from leading statistical journals. To construct StatEval, we develop \textbf{TRACE} (Topology and Reasoning-Aware Context Extractor), a multi-agent pipeline with human-in-the-loop validation that converts unstructured academic texts into self-contained theorem-level reasoning tasks. We also propose an Adaptive Process-Based Scoring Pipeline for complex statistical proofs, enabling fine-grained evaluation beyond final-answer matching. Experiments show that while LLMs perform reasonably on foundational tasks, they struggle with rigorous research-level reasoning. Beyond evaluation, StatEval serves as a resource for improving reasoning, as retrieval-augmented generation and domain-specific alignment consistently enhance performance. Together, these results establish StatEval as both a benchmark and an infrastructure for advancing statistical reasoning in LLMs.
\end{abstract}

\noindent%
{\it Keywords:} Large language model benchmark; Multi-agent data processing pipeline; Statistical reasoning; Retrieval-augmented generation; Domain-specific fine-tuning
\vfill

\newpage
\spacingset{1.8} 

\section{Introduction}
\label{sec:intro}
Statistics forms the foundation of modern data-driven science, with rigorous reasoning under uncertainty serving as its central pillar. Unlike purely computational tasks, statistical analysis fundamentally relies on sequential probabilistic reasoning to connect assumptions, models, data, and conclusions. Theoretically, one must verify regularity conditions, justify asymptotic approximations, and derive the properties of estimators and inferential procedures. In application, one must assess model fit, quantify uncertainty, and determine the robustness and generalizability of conclusions. Across both settings, every step of statistical analysis depends critically on careful probabilistic justification and logically coherent reasoning. Such reasoning is inherently complex, cognitively demanding, and highly susceptible to subtle errors that can yield misleading scientific conclusions or unreliable decisions.

Given the central role of reasoning in statistical analysis, recent advances in LLMs \citep{brown2020language, touvron2023llama} present unprecedented opportunities to assist with complex statistical reasoning and theoretical derivations. This potential is highlighted by rapid progress in mathematical reasoning \citep{guo2025deepseek} and automated proof discovery \citep{yu2025dapo}. However, enabling reliable AI assistance for statistics requires far more than increasingly capable models alone; it also depends critically on supporting infrastructure, including realistic benchmarks, curated corpora, and standardized evaluation protocols tailored to statistical reasoning. In other quantitative domains, such infrastructure has played a transformative role. Benchmarks such as GSM8K \citep{cobbe2021training} and MATH \citep{hendrycks2021math} have enabled systematic evaluation of mathematical reasoning, while HumanEval \citep{chen2021evaluating} has substantially accelerated progress in code generation and software engineering. In contrast, comparable infrastructure for statistics remains largely absent. Statistical problems constitute less than 3\% of recent reasoning benchmarks \citep{paster2024openwebmath}, and existing tasks are often restricted to isolated probability puzzles, lacking the layered inferential reasoning, asymptotic analysis, and uncertainty quantification that characterize real statistical practice \citep{gao2024omni}.

\begin{figure}[htbp]
    \centering
    \includegraphics[width=1\textwidth]{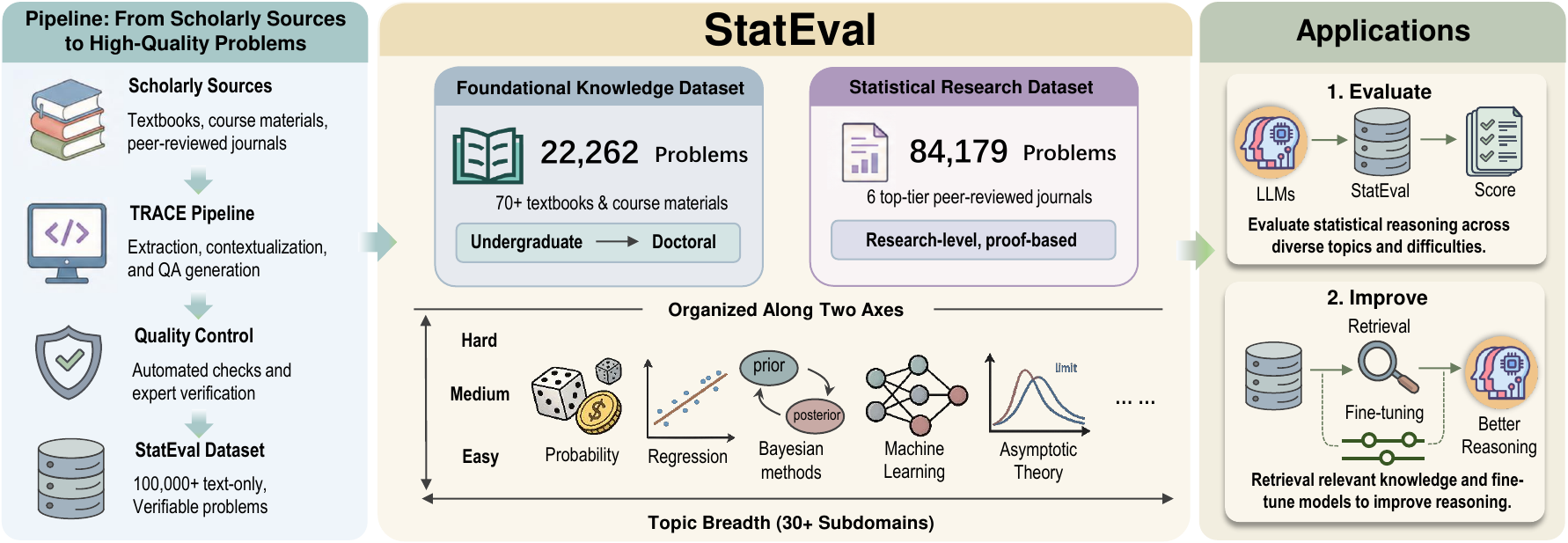}
    \caption{Overview of \textbf{StatEval}, illustrating the Foundational Knowledge and Advanced Statistical Research Datasets, with representative evaluations on core statistical reasoning tasks, including hypothesis testing, estimator asymptotics, and related inferential analysis.}
    \label{fig:StatEval_overview}
\end{figure}

To address this gap, we introduce \textbf{StatEval}, the first large-scale benchmark dedicated specially to statistical reasoning. \textbf{StatEval} is designed to span topics and difficulty levels in both breadth and depth, serving as a common resource for evaluation, RAG, and model adaptation. With over \textbf{100,000} meticulously curated items, \textbf{StatEval} covers the spectrum from introductory undergraduate exercises to advanced research-level challenges, capturing the diverse requirements of modern statistics (Figure~\ref{fig:StatEval_overview}). Because all problems are provided in a standardized, text-only form, \textbf{StatEval} can be readily integrated into a variety of workflows, including the evaluation of statistical logic, RAG pipelines, and model fine-tuning.


\textbf{StatEval} is organized along a difficulty axis that defines two complementary components:
\begin{itemize}[leftmargin=*]
    \item \emph{The foundational knowledge dataset} contains \textbf{22,262} problems drawn from over \textbf{70} textbooks and course materials spanning undergraduate to doctoral levels. It includes multiple-choice, short-answer, computational, and proof-based formats, and is intended to cover core statistical knowledge routinely taught in formal curricula.
    \item \emph{The statistical research dataset} consists of \textbf{84,179} verifiable, proof-based questions sourced from \textbf{6} top-tier peer-reviewed journals. These problems reflect research-oriented tasks that require rigorous derivations and proof-style solutions.
\end{itemize}
Within each component, we further introduce a disciplinary axis, organized as a two-level taxonomy that covers more than 30 subdomains, including probability, inference, regression, Bayesian methods, multivariate statistics, asymptotic theory, experimental design, and machine learning. This dual-axis design enables fine-grained slicing of problems by both difficulty and topic, supporting detailed analysis of model behavior as well as targeted retrieval or training in specific areas of statistics.

To construct \textbf{StatEval} at scale, we introduce \textbf{TRACE} (Topology and Reasoning-Aware Context Extractor), a multi-agent pipeline for extracting and organizing theorem-level reasoning data from statistical research papers. TRACE provides the infrastructure for converting unstructured academic texts into self-contained proof-based tasks, supporting both benchmark construction and model training. While TRACE is developed for \textbf{StatEval}, its modular design naturally generalizes to theorem-level data extraction in broader mathematical and scientific domains.

Using \textbf{StatEval} as a unified testbed, we develop a principled methodology for evaluating statistical reasoning in contemporary LLMs. To address the inherent complexity of mathematical proofs, we introduce an \emph{Adaptive Process-Based Scoring Pipeline}. Rather than relying on rigid reference matching, our framework dynamically accommodates multiple valid solution paths by employing granular, step-based verification for aligned strategies and independent logical validation for divergent approaches. This pipeline systematically synthesizes Logical Coherence, Technical Precision, and Terminal Accuracy. Applying this protocol to several state-of-the-art LLMs reveals that while they achieve reasonable performance on foundational statistical concepts and standard applied exercises, their capabilities remain substantially limited on research-level derivations and proofs. This performance gap underscores that the rigorous statistical reasoning abilities of current LLMs still have substantial room for improvement. 

Complementing this evaluation perspective, we then use the same \textbf{StatEval} corpus as a resource for \emph{improving} statistical reasoning via both retrieval and model adaptation. We design StatEval-based RAG systems in which models retrieve relevant problems, proofs, and solutions from the corpus to assist with new statistical tasks, and we examine the effects of fine-tuning on targeted subsets of \textbf{StatEval}, analyzing how exposure to foundational and research-level problems shapes subsequent reasoning performance. Across a range of benchmark tasks, these two intervention pathways—retrieval-augmented inference and supervised adaptation—yield consistent gains, particularly on proof-style questions. Together with the evaluation protocol, these results show how a single, carefully constructed dataset can simultaneously support rigorous measurement, targeted improvement, and more reliable deployment of LLMs for theoretical work in statistical practice. 

Our main contributions are as follows:
\begin{enumerate}[leftmargin=*]
    \item First, we introduce a large-scale statistical reasoning benchmark comprising over 100,000 curated problems spanning both foundational and research-level settings. StatEval fills a critical gap in existing reasoning benchmarks, which largely lack dedicated resources for rigorous and comprehensive statistical reasoning tasks.
    \item Second, we develop \textbf{TRACE}, an automated multi-agent pipeline for large-scale theorem-level problem extraction and logical parsing from unstructured academic texts. Beyond statistics, TRACE provides a generalizable framework for constructing reasoning datasets in other domains requiring rigorous multi-step derivations.
    \item Third, we propose an evaluation framework for complex reasoning tasks that goes beyond final-answer matching by assessing logical coherence, technical correctness, and process completeness step-by-step. This protocol offers a scalable and flexible foundation for evaluating proof-based reasoning in mathematics and related disciplines.
    \item Fourth, we demonstrate that \textbf{StatEval} functions not only as a benchmark, but also as a resource for improving statistical intelligence in LLMs. Through empirical studies, we show that leveraging StatEval via retrieval-augmented generation and domain-specific fine-tuning consistently strengthens rigorous statistical reasoning capabilities.
\end{enumerate}

The remainder of this paper is organized as follows. 
Section~\ref{sec:stateval} introduces the \textbf{StatEval} dataset and its distributional characteristics. 
Section~\ref{sec:pipeline} describes the multi-stage data processing pipeline for benchmark construction. 
Sections~\ref{sec:app_benchmark}, \ref{sec:rag_app}, and \ref{sec:finetuning_app} collectively demonstrate the practical utility of our dataset: evaluating contemporary LLMs via an adaptive scoring protocol, exploring RAG-based augmentation to enhance multi-step reasoning, and investigating domain-specific fine-tuning to align open-weights models with rigorous statistical derivations. 
Finally, Section~\ref{sec:conclusion} concludes the paper.

\section{Breakdown of StatEval}
\label{sec:stateval}

This section systematically describes a systematic description of \textbf{StatEval}, a benchmark spanning foundational and frontier statistical topics. 
\textbf{StatEval} is structured along the difficulty axis into two parts: foundational knowledge and statistical research dataset. 
For each dataset, we detail its data sources, question formats, and disciplinary structure, which support fine-grained evaluation across a wide spectrum of statistical subdomains. 

\subsection{Foundational Knowledge Dataset}
\label{sec:fundamental}

The foundational knowledge dataset evaluates LLMs' mastery of statistical concepts and ability to solve classic problems through rigorous reasoning. Covering both undergraduate foundations and graduate-level training, it provides a basis for assessing academic proficiency. 

\paragraph*{Scale and Data Sources.}
The dataset comprises \textbf{22,262} problems, stratified by academic difficulty, including 9,382 undergraduate-level and 12,880 graduate-level instances.
The problems are curated from three primary sources to ensure pedagogical validity and coverage: (i) 76 classical textbooks in statistics and related fields, which constitute the core curriculum; (ii) over one thousand carefully verified, exam-style questions from graduate entrance examinations and exercise collections, establishing a rigorous testing benchmark; and (iii) selected problems from publicly available courses at leading international universities, supplemented by high-quality online resources to enhance topical diversity.

\paragraph*{Question Formats.}
To rigorously evaluate different cognitive capabilities, the dataset features five distinct problem formats: formal proofs, calculation-intensive questions, short-answer inquiries, multiple-choice questions, and fill-in-the-blank exercises. The distribution is intentionally heavily weighted toward open-ended generative tasks---specifically over 10,000 proof-based problems and nearly 6,000 calculation questions---to shift the evaluation focus from simple pattern matching to deep logical derivation and structural problem-solving.

\begin{figure}[htbp]
    \centering
    \includegraphics[width=1.0\textwidth]{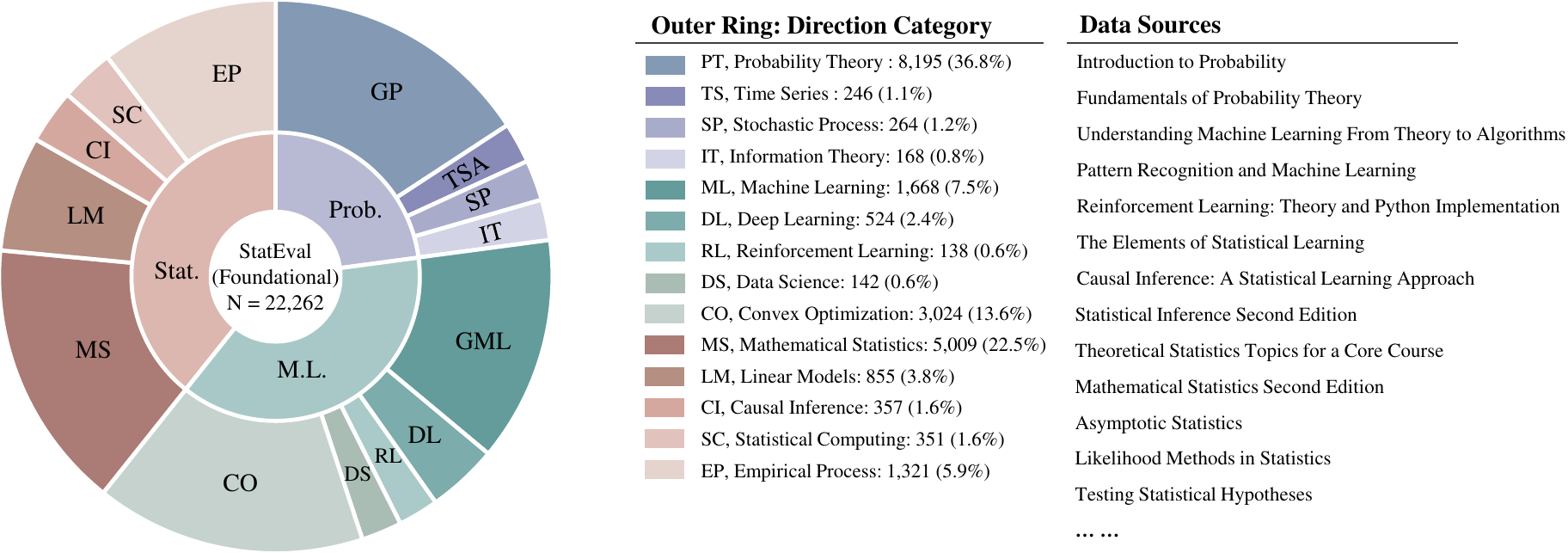} 
    \caption{Disciplinary classification of foundational-level datasets}
    \label{fig:foundational_composition} 
\end{figure}

\paragraph*{Disciplinary Structure.}
Content is organized hierarchically to facilitate granular analysis.
At the primary tier, problems are grouped into three domains: Probability, Statistics, and Machine Learning. 
At the second tier, each domain is divided into course-level subjects with distinct undergraduate and graduate scopes. 
For instance, undergraduate subjects primarily focus on foundational disciplines such as Probability, Mathematical Statistics, Stochastic Processes, and Machine Learning, whereas the graduate level extends to advanced fields including Measure Theory, Empirical Processes, and Reinforcement Learning. 
The relative proportions of these categories, alongside representative source materials, are visualized in Figure~\ref{fig:foundational_composition}.

\subsection{Statistical Research Dataset}
\label{sec:research}

The statistical research dataset is designed to benchmark the ability of LLMs to perform structured, multi-step reasoning on frontier research-level problems. Unlike foundational tasks, these problems require models to navigate complex theoretical landscapes, bridging precise assumptions with rigorous derivations to establish novel statistical guarantees.

\paragraph*{Scale and Data Sources.}
This dataset represents a significant scaling of formal statistical evaluation, comprising \textbf{84,179} proof-based research tasks derived from \textbf{6,953} high-impact research articles published between \textbf{2000 and 2025}. Based on whether prerequisite theorems are provided, these problems are categorized into three hierarchical difficulty levels: \textbf{40,366} instances at the \textit{Easy} level, \textbf{22,013} at \textit{Medium}, and \textbf{21,800} at \textit{Hard}. The corpus is primarily sourced from six top-tier journals: \textit{Annals of Statistics}, \textit{Journal of the American Statistical Association}, \textit{Journal of the Royal Statistical Society: Series B}, \textit{Biometrika}, \textit{Statistica Sinica}, and the \textit{Journal of Machine Learning Research}.

\paragraph*{Question Formats.}
To comprehensively evaluate the nuanced statistical reasoning capabilities of LLMs, we standardize the problem formulation based on the inherent logical structures of the research papers. Since mathematical theorems frequently depend on prerequisite results, we leverage the topological dependency graph and the availability of prerequisite theorems to construct three progressive difficulty levels:
\begin{itemize}[leftmargin=*]
    \item \textbf{Easy difficulty:} Prerequisite theorems or lemmas are explicitly provided in the problem prompt as given facts. The model is only required to execute the final proof of the target theorem using these provided conditions.
    \item \textbf{Medium difficulty:} The contents of the prerequisite theorems are provided in the prompt, but the model is required to formally prove these prerequisites step-by-step before deriving the main target theorem.
    \item \textbf{Hard difficulty:} No prerequisite hints or theorems are provided. The model must independently construct the entire logical proof chain, formulating the necessary intermediate lemmas from scratch to reach the final conclusion.
\end{itemize}

\begin{figure}[htbp]
  \centering
  \includegraphics[width=\textwidth]{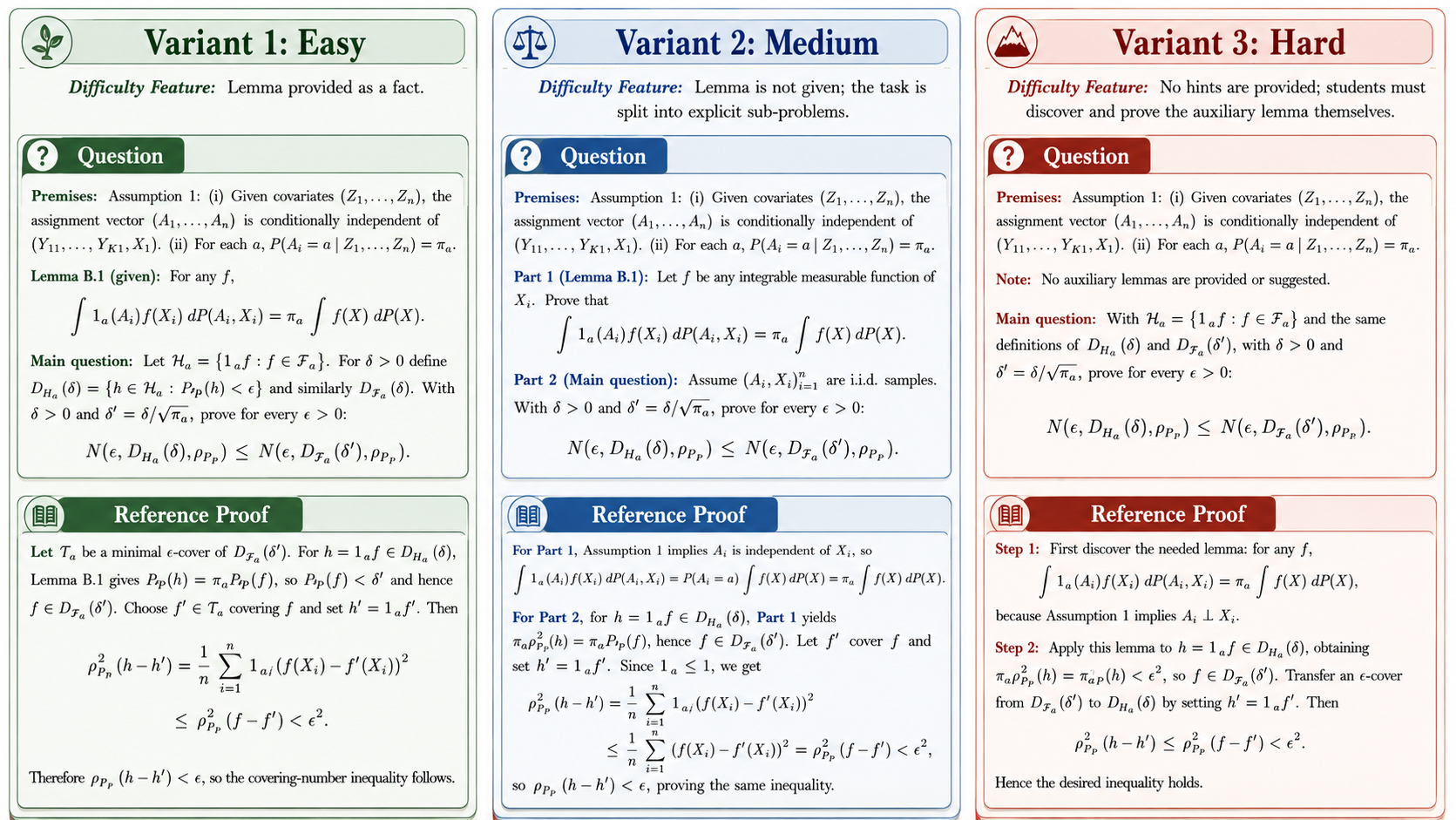}
  \caption{Representative research-level problem variants, illustrating the progression from Easy to Medium and Hard settings.}
  \label{fig:research_example}
\end{figure}

Figure~\ref{fig:research_example} presents a representative theorem instantiated under the Easy, Medium, and Hard settings. For readability, the displayed derivations are truncated while preserving the key structural differences across difficulty levels.

\begin{figure}[htbp]
    \centering
    \includegraphics[width=1.0\textwidth]{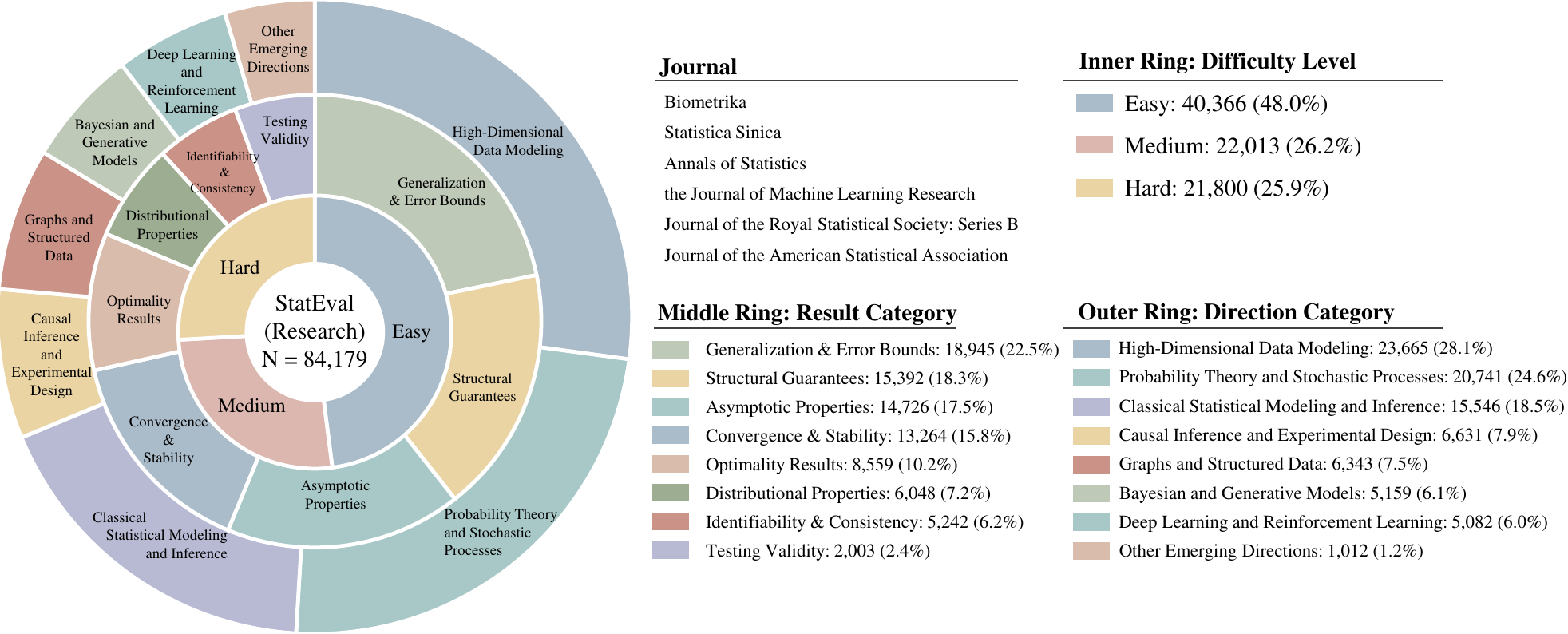} 
    \caption{Disciplinary classification of statistical research datasets.}
    \label{fig:research_composition}  
\end{figure}

\paragraph*{Disciplinary and Theoretical Structure.}
To support fine-grained diagnosis of reasoning capabilities, the dataset is structured along two orthogonal axes: \textit{Research Subfield} and \textit{Theoretical Property}. Subfields such as High-dimensional Data Modeling, Causal Inference, and Deep Learning Theory define the domain context, while the \textbf{Theoretical Property} axis classifies the mathematical nature of the result. To efficiently and consistently annotate the 84,179 theorems, we use an advanced LLM within our data processing pipeline to automatically classify each task across both dimensions. Based on this taxonomy, we categorize the theoretical properties into eight result types: \textbf{Asymptotic Properties}, \textbf{Convergence \& Stability}, \textbf{Distributional Properties}, \textbf{Generalization \& Error Bounds}, \textbf{Identifiability \& Consistency}, \textbf{Optimality Results}, \textbf{Structural Guarantees}, and \textbf{Testing Validity}. This stratification enables StatEval to pinpoint specific weaknesses in a model's theoretical reasoning capabilities. The detailed distribution of problems across research subfields is visualized in Figure~\ref{fig:research_composition}. Additional dataset specifications, including detailed distributions and quality control procedures, are provided in the supplementary material.

\section{Data Processing Pipeline: TRACE}
\label{sec:pipeline}

\begin{figure}[htbp]
    \centering
    \includegraphics[width=1.0\textwidth]{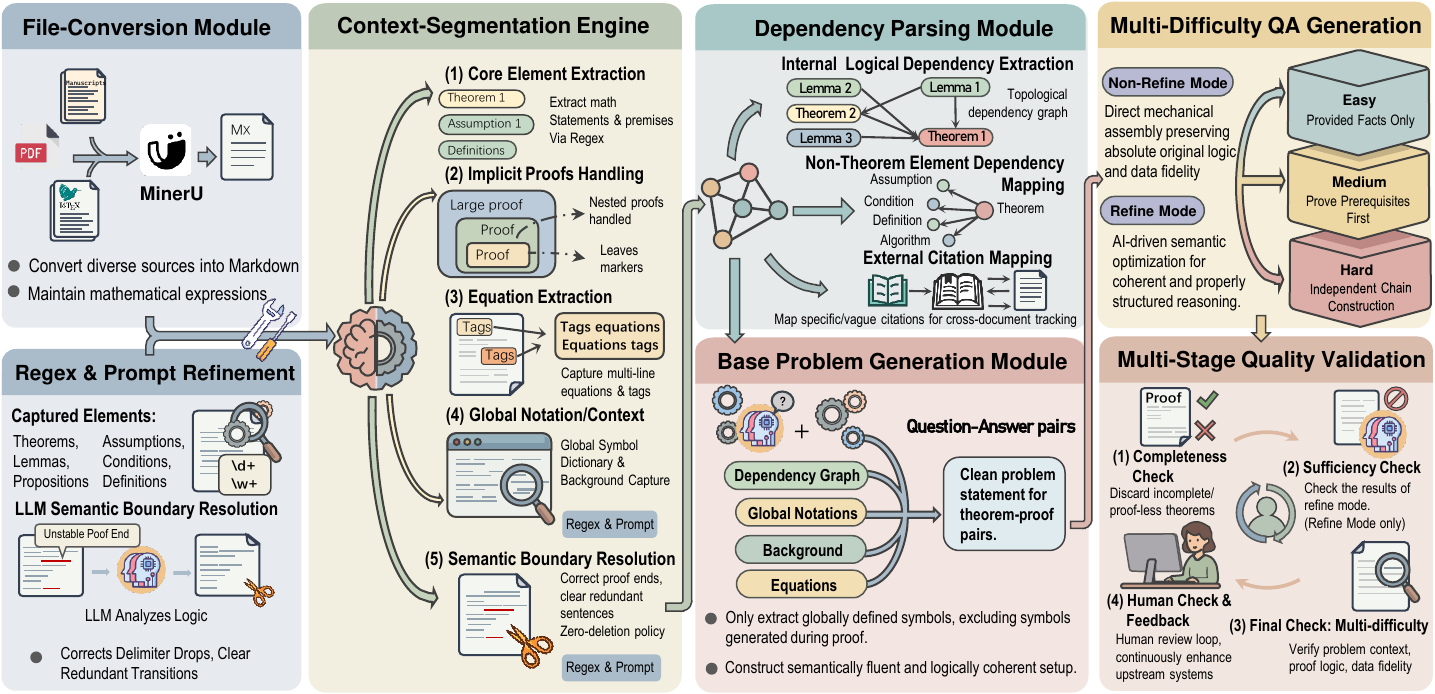}
    \caption{Overview of the TRACE pipeline. Each module corresponds to a major functional stage in automated document conversion, contextual segmentation, dependency parsing, multi-level question generation, and quality validation.}
    \label{fig:data_pipeline}
\end{figure}

In this section, we introduce \textbf{TRACE}, a multi-agent data processing pipeline for constructing high-fidelity statistical reasoning corpora from heterogeneous research documents. As illustrated in Figure~\ref{fig:data_pipeline}, TRACE consists of six functional modules: the File-Conversion Module, the Context-Segmentation Engine, the Base Problem Generation Module, the Dependency Parsing Module, the Multi-Difficulty Question Generation Module, and the Multi-Stage Quality Validation Module. These modules jointly support two complementary goals: (i) extracting reliable theorem-proof corpora from unstructured statistical literature at scale, and (ii) transforming them into systematic problems for evaluating and improving the mathematical reasoning capabilities of LLMs.

The design of TRACE addresses a critical tension in parsing advanced statistical literature: the trade-off between deterministic structural extraction and semantic generalization. LLM-only approaches often exhibit symbolic hallucinations, truncation of complex derivations, and systemic omission of critical prerequisite lemmas \citep{zhang2025realmath, blecher2023nougat}. Conversely, rule-based methods are brittle, especially under PDF conversion artifacts, irregular theorem environments, and cross-paragraph contextual dependencies \citep{lopez2009grobid}. TRACE therefore combines robust regular-expression engines for structural extraction with LLM agents for semantic orchestration, proof reconstruction, and quality control.

\subsection{File Conversion and Context-Segmentation Engine}
\label{sec:parsing_decoupling}

TRACE first standardizes raw research documents through the File-Conversion Module, which converts heterogeneous PDF sources into structured Markdown using MinerU for formula-aware optical character recognition. This step preserves mathematical expressions, theorem environments, section boundaries, and local textual context, providing a normalized representation for downstream parsing.

The Context-Segmentation Engine then isolates core theoretical elements, including theorems, lemmas, assumptions, definitions, equations, and proofs. Instead of relying on a single global parser, TRACE employs layered extraction: explicit theorem-proof blocks are captured directly, implicit proof regions are recovered via local structural cues, and equation environments are separately indexed for later reference. The engine also inserts structural markers into the extracted text, representing nested or distributed proof components as independent data objects while preserving their original hierarchical relations.

\begin{figure}[htbp]
    \centering
    \includegraphics[width=\textwidth]{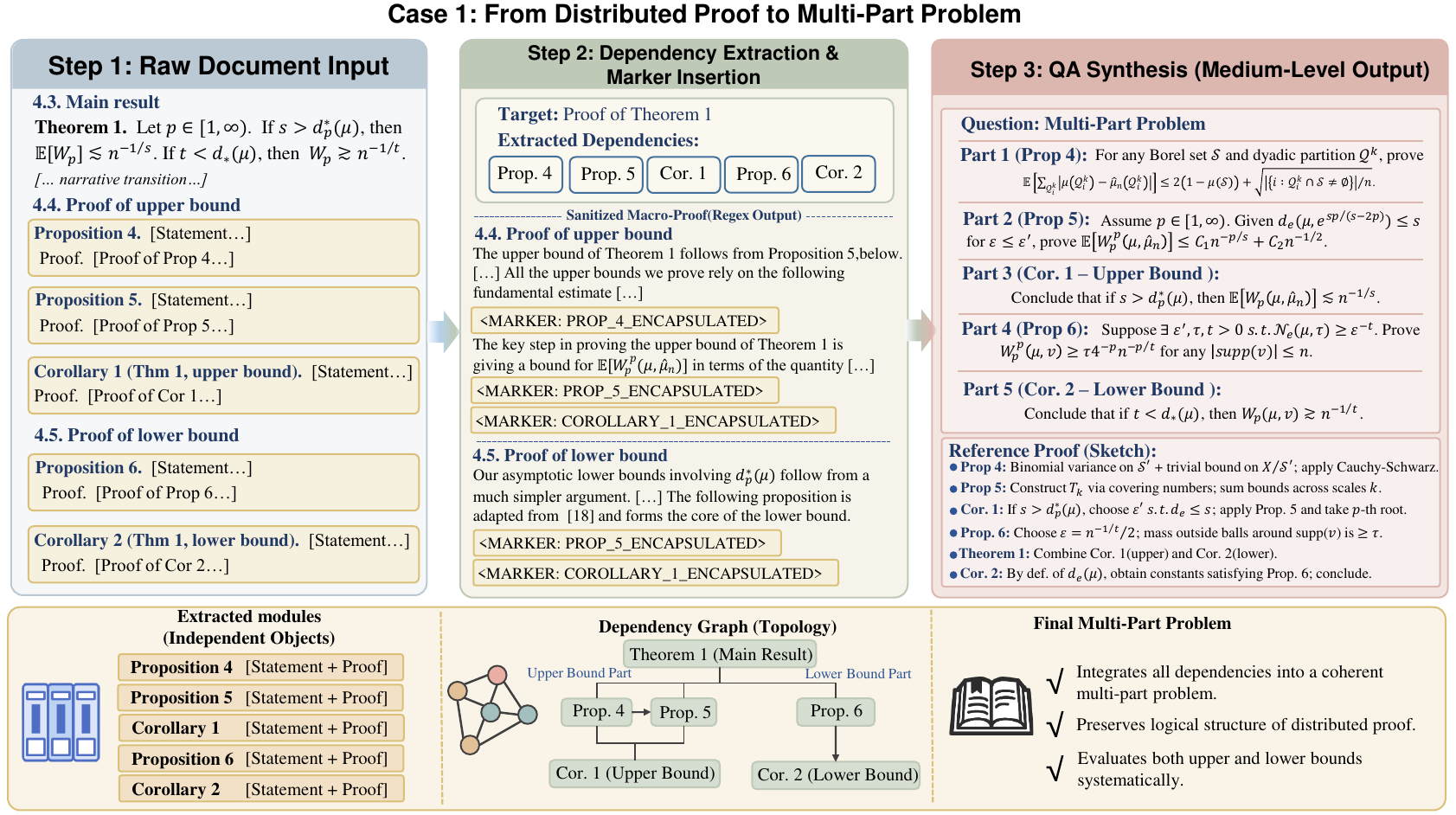}
    \caption{Illustration of the TRACE pipeline processing a distributed proof, using Theorem 1 of \citet{weed2019sharp}, whose proof is split across upper (Section 4.4) and lower  (Section 4.5) bounds.}
    \label{fig:case1}
\end{figure}

This capability is crucial for handling distributed and nested proofs, which are common in statistical papers. A main theorem may rely on auxiliary propositions, intermediate claims, or separate upper- and lower-bound arguments proved in different sections. As illustrated in Figure~\ref{fig:case1}, TRACE decouples these intertwined components, records their relations in a unified structural representation, and replaces embedded proof fragments with explicit markers. In later stages, these markers allow LLM agents to reconstruct a self-contained problem while retaining the macro-logical structure of the original paper.

\subsection{Notation Harmonization and Base Problem Generation}
\label{sec:context_harmonization}

After structural segmentation, TRACE further ensures that each extracted theorem remains self-contained. Statistical papers frequently define key parameters, operators, estimators, and regularity conditions outside the immediate theorem environment. Directly extracting a theorem-proof block without this background context may cause notation ambiguity or missing assumptions. To address this issue, TRACE constructs a document-level context pool by combining local sliding-window scans, globally declared notation sections, surrounding assumptions, and indexed equation references. An LLM agent then consolidates this information into a validated Global Notation Reference Table.

\begin{figure}[htbp]
    \centering
    \includegraphics[width=\textwidth]{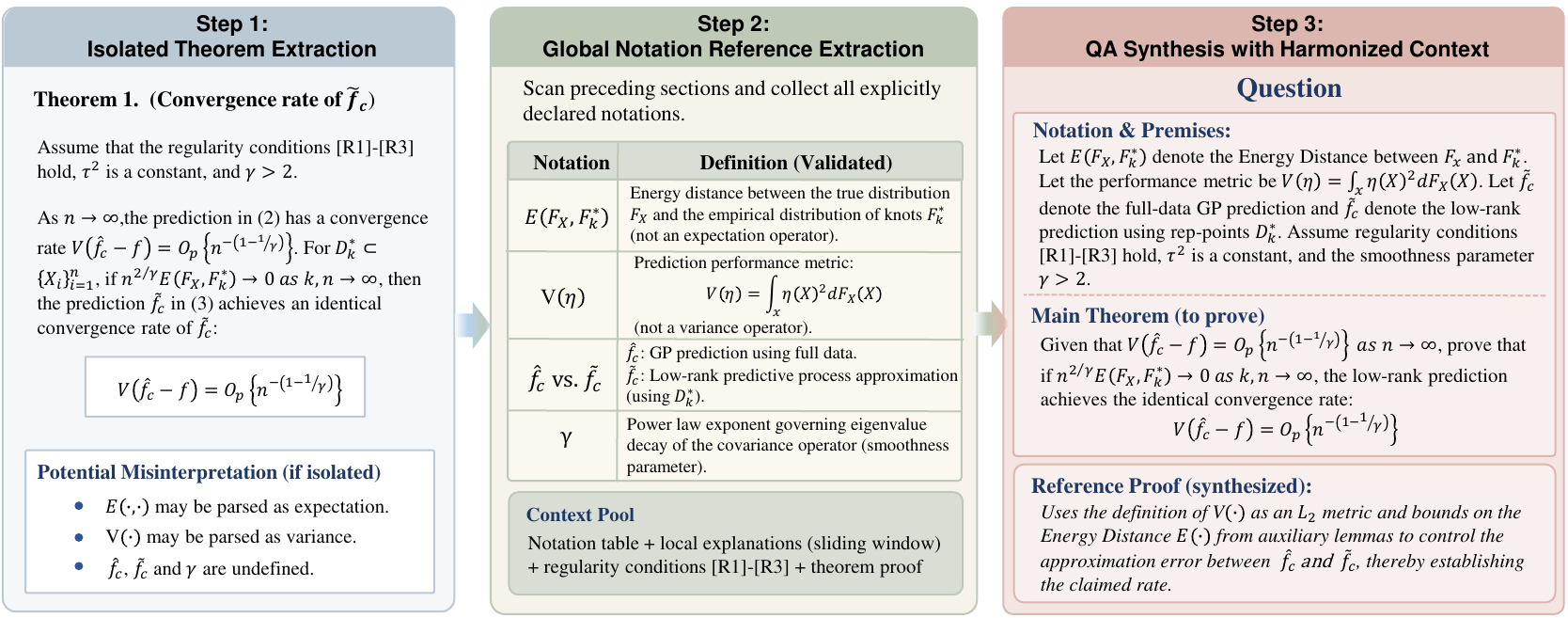}
    \caption{Illustration of TRACE addressing cross-paragraph notation harmonization to prevent semantic shifts, using Theorem 1 of \citet{song2025large} on large-scale low-rank Gaussian process prediction.}
    \label{fig:case2}
\end{figure}

This notation harmonization step prevents semantic shifts caused by isolated extraction. As shown in Figure~\ref{fig:case2}, symbols such as $E(\cdot,\cdot)$ and $V(\cdot)$ may be incorrectly interpreted as expectation or variance operators when read out of context, while important parameters may remain undefined. By explicitly incorporating document-specific notation and background assumptions into the problem setup, TRACE ensures that the resulting benchmark evaluates statistical reasoning rather than a model's ability to infer missing definitions.

The Base Problem Generation Module then assembles the segmented theorem, proof, background assumptions, global notations, and relevant equations into a clean theorem-proof pair. At this stage, TRACE aims to produce a faithful base problem that contains all necessary definitions while excluding unrelated symbols and redundant context. The output serves as the canonical problem unit for subsequent dependency expansion and multi-difficulty Question generation.

\subsection{Dependency Parsing and Multi-Difficulty Question Synthesis}
\label{sec:dependency_mapping}

Given the base problem, TRACE further constructs a comprehensive dependency graph through the Dependency Parsing Module. This module first detects candidate links among local labels, prerequisite theorems, assumptions, definitions, equations, and external citations using enhanced regular-expression matching. Because not every textual citation corresponds to a genuine deductive dependency, TRACE applies an LLM-based semantic filtering layer to remove superficial references, such as citations used only for notation borrowing. The remaining dependencies are organized into a topological graph that specifies which theoretical elements must be provided to make a problem logically solvable.

Using this dependency graph, the Multi-Difficulty Question Generation Module synthesizes Easy, Medium, and Hard variants under two modes:
\begin{itemize}[leftmargin=*]
    \item \textbf{Non-Refine Mode:} This mode prioritizes textual fidelity by mechanically concatenating the base problem with the required prerequisite lemmas or definitions. Since it avoids LLM-driven rewriting, it minimizes hallucination risk and produces high-fidelity data suitable for RAG-style retrieval and model fine-tuning.
    \item \textbf{Refine Mode:} This mode target evaluation-quality problem construction. It applies a \textit{Context-Aware Patching} mechanism in which an LLM agent expands skipped intermediate steps, resolves cross-theorem logical jumps, and corrects local typographical artifacts. All edits follow a strict \textit{Zero-Deletion Policy}: the model may only perform localized additive revisions and cannot remove or rewrite the original derivation.
\end{itemize}

\begin{figure}[htbp]
    \centering
    \includegraphics[width=\textwidth]{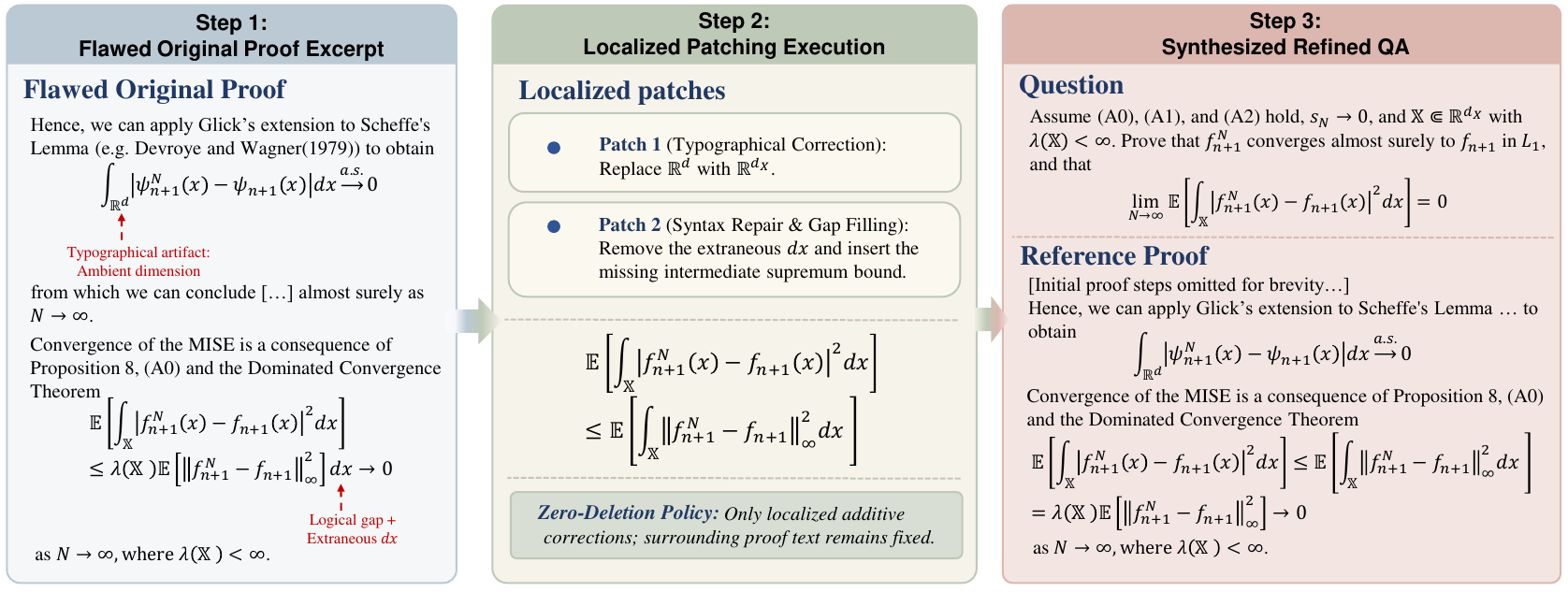}
    \caption{Illustration of TRACE executing Context-Aware Patching to resolve typographical artifacts and logical jumps via localized expansions, demonstrated using an excerpt establishing the convergence of the Mean Integrated Squared Error (MISE) from \citet{crucinio2023particle}.}
    \label{fig:case3}
\end{figure}

The distinction between these two modes is important because evaluation data must be both faithful to the original proof and sufficiently explicit for rigorous assessment. Unconstrained proof rewriting can erase valid author-specific techniques or oversimplify derivations. As illustrated in Figure~\ref{fig:case3}, TRACE instead identifies localized anomalies, defines bounded context windows around them, and patches only the affected regions---for example, by correcting notation or inserting missing intermediate inequalities. This preserves the original proof strategy while making the final problem-proof pair logically complete and readable.

Finally, TRACE applies a Multi-Stage Quality Validation Module to ensure the reliability of the generated corpus. A completeness check first removes samples with severe OCR truncation, missing theorem statements, or broken proof structures. For refined samples, a sufficiency check verifies that the problem setup contains all assumptions, global notations, and prerequisite results needed by the proof. A consistency check then confirms that the generated Easy, Medium, and Hard variants satisfy their intended dependency constraints. Finally, a subset of samples undergoes human expert review, whose feedback is used to refine extraction rules, improve prompting templates, and further suppress hallucinated or logically inconsistent outputs.

\section{Application I: Comprehensive Evaluation of Statistical Reasoning}
\label{sec:app_benchmark}

To assess the intrinsic statistical reasoning capabilities of current LLMs, we conduct a comprehensive benchmark evaluation on \textbf{StatEval-mini}, a compact evaluation subset sampled from the full StatEval benchmark and publicly released on Hugging Face (\url{https://huggingface.co/datasets/StatAILab/StatEval}). This section presents the benchmark construction, model coverage, scoring protocol, and performance analysis across both foundational and research-level statistical reasoning tasks.

\subsection{Experimental Setup}
\paragraph*{Dataset Construction.}
To balance computational cost with statistical coverage, we construct a compact yet representative evaluation subset, \textbf{StatEval-mini}, drawn from both the Foundational Knowledge and Statistical Research datasets. From the foundational dataset of 22,262 problems, we sample 1,000 representative questions stratified by statistical sub-discipline and academic level. From the research dataset of 84,179 tasks, we select 900 instances corresponding to the Easy, Medium, and Hard variants of 300 unique theorems with explicit dependency structures. These research instances are further stratified by journal, statistical sub-discipline, and theoretical property.

To avoid contamination and evaluate reasoning on recent unseen developments, all research-level test instances are drawn from 2024--2025 articles. The remaining data are used as the retrieval corpus for RAG and the fine-tuning training corpus, see Sections~\ref{sec:rag_app} and~\ref{sec:finetuning_app}. 


\paragraph*{Evaluated Models.}
We evaluate a diverse set of frontier language models, categorizing them into proprietary and open-weight architectures. The proprietary models include GPT-5.4 developed by OpenAI, Gemini-3.1-Pro by Google, Claude-Opus-4.7 by Anthropic, Qwen-3.6-Plus by Alibaba, and Doubao-Seed-2.0-Pro by ByteDance. For open-weight models, we consider Llama-4-Maverick-17B developed by Meta, DeepSeek-V3.2-Thinking (specifically designed for reasoning) and DeepSeek-V4-Pro, both developed by DeepSeek.

\subsection{Evaluation Protocol}
\label{sec:scoring_protocol}

To ensure rigorous and interpretable evaluation, \textbf{StatEval} adopts scoring mechanisms tailored to different question types. Multiple-choice questions are evaluated by exact matching, while open-ended derivations and proofs are assessed using an adaptive process-based scoring pipeline. We use \textbf{GPT-5.4-mini} as the automatic judge under this protocol.

\begin{figure}[htbp]
    \centering
    \includegraphics[width=0.9\textwidth]{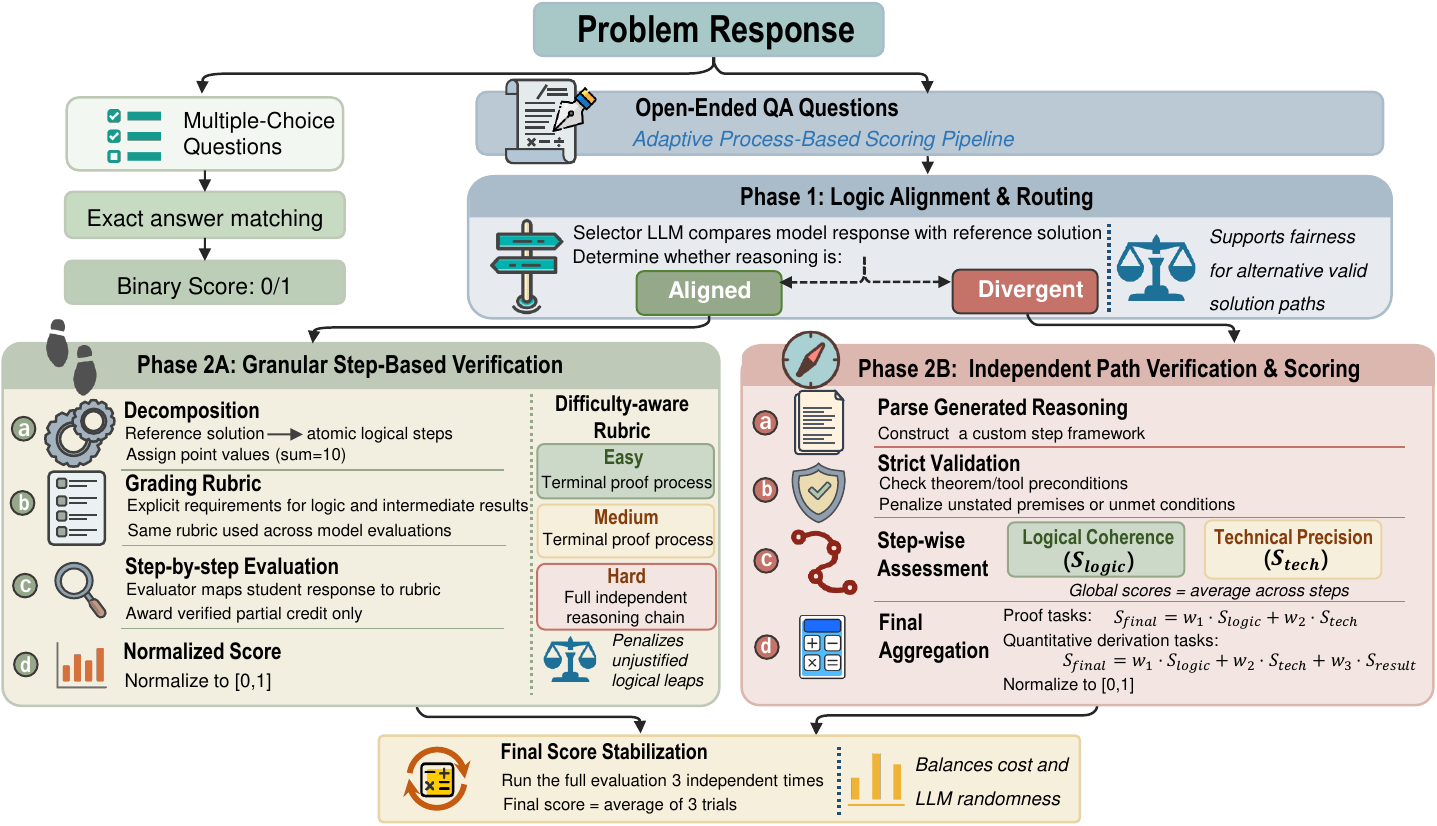} 
    \caption{Adaptive scoring pipeline in StatEval. Responses are routed to either reference-based step verification or independent path verification according to their alignment with the reference solution.}
    \label{fig:eval_process}  
\end{figure}

\paragraph*{Scoring for Multiple-Choice Questions.}
Multiple-choice questions are graded by exact answer matching, resulting in a binary score of 0 or 1.

\paragraph*{Scoring for Open-Ended Questions.}
For open-ended questions, we introduce an \texttt{Adaptive Process-Based Scoring Pipeline}. As shown in Figure~\ref{fig:eval_process}, the pipeline first determines whether the model follows the reference reasoning path and then applies the corresponding scoring branch.

\begin{enumerate}[leftmargin=*]
    \item \textbf{Phase 1: Logic Alignment and Routing.}
    
    Since mathematical problems may admit multiple valid solution paths, strict comparison with a single reference solution can unfairly penalize alternative derivations. A Selector LLM therefore compares the model response with the reference solution at the level of proof strategy and routes it as either \textbf{Aligned} or \textbf{Divergent}.

    \item \textbf{Phase 2A: Reference-Based Step Verification.}

    For \textbf{Aligned} responses, the reference solution is decomposed into atomic logical reasoning steps to form a grading rubric. An independent evaluator then maps the model response onto this rubric and assigns partial credit according to correctly executed components. The rubric is adapted to each problem's difficulty, allowing the score to reflect genuine logical progress while penalizing unjustified leaps. The final score is normalized to $[0,1]$.

    \item \textbf{Phase 2B: Independent Path Verification.}

    For \textbf{Divergent} responses, the system evaluates the alternative reasoning path independently rather than forcing it onto the reference solution. The evaluator first checks whether the invoked assumptions, theorem conditions, and theoretical boundaries are valid. It then scores the response along two dimensions:
    \begin{itemize}[leftmargin=*]
        \item \textbf{Logical Coherence ($S_{\text{logic}}$):} validity and consistency of the overall proof strategy and step-to-step inferential transitions.
        \item \textbf{Technical Precision ($S_{\text{tech}}$):} correctness and rigor of algebraic, probabilistic, and analytical derivations.
    \end{itemize}

    For proof-based tasks, the final score is computed as
    \[
    S_{\text{final}} = w_1 S_{\text{logic}} + w_2 S_{\text{tech}}.
    \]
    For non-proof quantitative derivations, terminal accuracy is additionally considered:
    \[
    S_{\text{final}} = w_1 S_{\text{logic}} + w_2 S_{\text{tech}} + w_3 S_{\text{result}}.
    \]
\end{enumerate}

\begin{figure}[t]
    \centering
    \includegraphics[width=\linewidth]{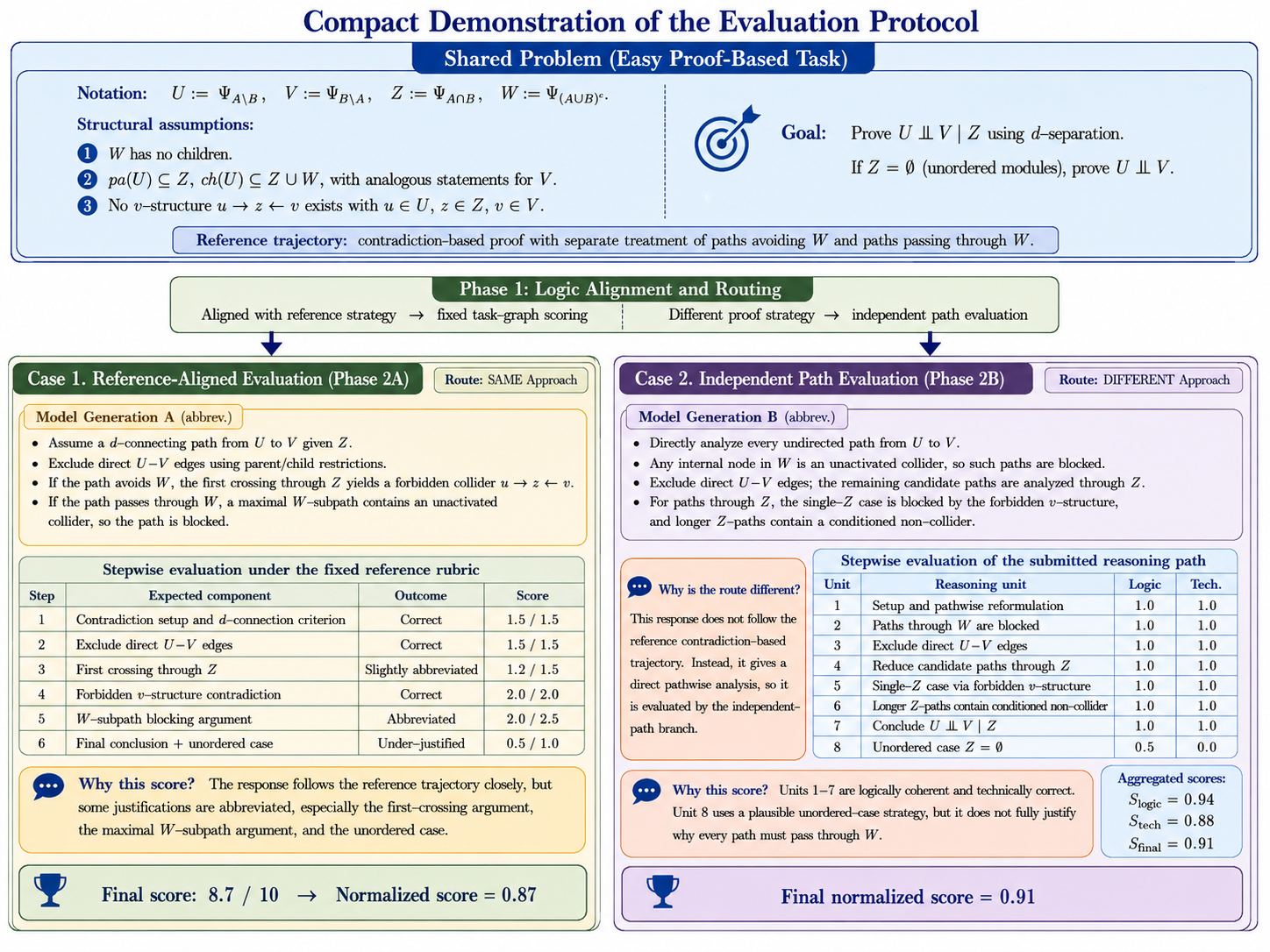}
    \caption{
    Compact demonstration of the evaluation protocol on an easy proof-based task adapted from Lemma 3 of ~\cite{liu2025generalframework}.
    }
    \label{fig:evaluation_case}
\end{figure}

Figure~\ref{fig:evaluation_case} illustrates the two distinct scoring branches using a single proof task as an example. In Case 1, the response follows the reference trajectory and is scored by reference-based step verification. In Case 2, the response uses an alternative method and is therefore evaluated through independent path verification, where the proof is decomposed into reasoning units and scored by logical coherence and technical precision.

To reduce evaluation variance, each response is judged three independent times, and the final score is computed as the average across trials. Additional details, including complete performance breakdowns, routing statistics, multi-judge validation, and protocol diagnostics, are provided in the supplementary material.

\subsection{Evaluation Results}

We report the main evaluation results under the above protocol. The final leader board score is computed as the total points obtained divided by the total possible points.

\paragraph*{Results for the Foundational Knowledge Dataset.} 

Table~\ref{tab:model_results_final} summarizes the performance across academic levels (undergraduate and graduate) and three major domains: probability, statistics, and machine learning.

\begin{table}[htbp]
\centering
\small
\renewcommand{\arraystretch}{1.0}
\setlength{\tabcolsep}{4pt}
\caption{\textbf{Foundational knowledge dataset results by academic level and domain.}}
\label{tab:model_results_final}
\begin{tabular}{lccccccccc}
\toprule
\multirow{2}{*}{\textbf{Model}} & \multicolumn{4}{c}{\textbf{Graduate}} & \multicolumn{4}{c}{\textbf{Undergraduate}} & \textbf{Overall} \\
\cmidrule(lr){2-5} \cmidrule(lr){6-9} \cmidrule(lr){10-10}
& \textbf{Prob.} & \textbf{Stat.} & \textbf{ML} & \textbf{Mean} & \textbf{Prob.} & \textbf{Stat.} & \textbf{ML} & \textbf{Mean} & \textbf{Mean} \\
\midrule
\multicolumn{10}{l}{\textit{Proprietary Models}} \\
Gemini-3.1-Pro & 89.07 & 87.78 & 92.60 & 89.07 & \textbf{94.59} & \textbf{89.86} & 91.13 & \textbf{92.26} & \textbf{90.83} \\
Qwen-3.6-Plus & \textbf{93.05} & 89.42 & 88.46 & \textbf{90.15} & 91.52 & 87.24 & 87.74 & 89.29 & 89.68 \\
GPT-5.4 & 86.31 & 82.76 & 90.36 & 85.19 & 90.59 & 86.86 & \textbf{92.04} & 89.48 & 87.55 \\
Claude-Opus-4.7 & 90.94 & 86.39 & 83.29 & 86.92 & 88.90 & 89.47 & 79.46 & 87.48 & 87.23 \\
Doubao-Seed-2.0-Pro & 84.71 & 84.40 & 83.23 & 84.24 & 87.69 & 82.02 & 84.69 & 85.10 & 84.71 \\
\midrule
\multicolumn{10}{l}{\textit{Open-Weight Models}} \\
DeepSeek-V4-Pro & 81.95 & \textbf{91.78} & \textbf{94.94} & 89.65 & 80.19 & 88.68 & 76.31 & 82.92 & 90.66 \\
DeepSeek-V3.2-thinking & 87.84 & 86.95 & 88.21 & 87.42 & 90.18 & 85.78 & 79.80 & 86.78 & 87.07 \\
Llama-4-Maverick-17B & 75.59 & 74.13 & 75.09 & 74.69 & 75.48 & 73.68 & 78.52 & 75.35 & 75.05 \\
\bottomrule
\end{tabular}
\end{table}


As shown in Table~\ref{tab:model_results_final}, frontier models exhibit near-saturated performance on the Foundational Knowledge Dataset, with the top models clustered around 90\% overall accuracy. The results also reveal domain-specific strengths: DeepSeek-V4-Pro performs particularly well on graduate-level Statistics and Machine Learning, whereas Gemini-3.1-Pro leads on undergraduate-level domains. Even mid-tier and smaller models retain substantial baseline competence, mostly exceeding 70\%. These results suggest that textbook-level statistical concepts and routine derivations are already well internalized by current LLMs, leaving limited headroom for differentiation on foundational tasks.

\paragraph*{Results for the Statistical Research Dataset.} 

Table~\ref{tab:research_results} summarizes the models' statistical reasoning performance across the three hierarchical difficulty levels on the frontier research tasks.

\begin{table}[htbp]
\centering
\small
\renewcommand{\arraystretch}{1.0}
\setlength{\tabcolsep}{12pt}
\caption{\textbf{Statistical research dataset results by difficulty level.}}
\label{tab:research_results}
\begin{tabular}{l c c c c}
\toprule
\multirow{2}{*}{\textbf{Model}} & \multicolumn{3}{c}{\textbf{Difficulty}} & \textbf{Overall} \\
\cmidrule(lr){2-4}
& \textbf{Easy} & \textbf{Medium} & \textbf{Hard} & \textbf{Mean} \\
\midrule
\multicolumn{5}{l}{\textit{Proprietary Models}} \\
GPT-5.4 & \textbf{68.28} & \textbf{65.57} & \textbf{51.28} & \textbf{61.70} \\
Claude-Opus-4.7 & 56.79 & 54.25 & 42.02 & 51.02 \\
Qwen-3.6-Plus & 55.42 & 51.92 & 40.05 & 49.14 \\
Gemini-3.1-Pro & 55.55 & 49.18 & 37.02 & 47.25 \\
Doubao-Seed-2.0-Pro & 47.22 & 44.31 & 33.55 & 41.69 \\
\midrule
\multicolumn{5}{l}{\textit{Open-Weight Models}} \\
DeepSeek-V4-Pro & 65.73 & 57.81 & 41.86 & 55.15 \\
DeepSeek-V3.2-thinking & 56.64 & 51.43 & 38.72 & 48.93 \\
Llama-4-Maverick-17B & 38.29 & 27.06 & 22.43 & 29.26 \\
\bottomrule
\end{tabular}
\end{table}


As shown in Table~\ref{tab:research_results}, model performance declines monotonically as difficulty increases, revealing a sharp contrast with the foundational evaluation. Models that approach 90\% accuracy on textbook-level tasks suffer substantial drops on frontier research proofs. GPT-5.4 is the most robust model, maintaining over 50\% even on Hard tasks, but the overall gap remains large. This suggests that current LLMs are far better at applying familiar curriculum-level patterns than at performing long-horizon, out-of-distribution theoretical reasoning. The pronounced degradation of smaller models further indicates that advanced statistical reasoning remains strongly scale-dependent.

\section{Application II: Retrieval-Augmented Generation}
\label{sec:rag_app}

\begin{figure}[htbp]
    \centering
    \includegraphics[width=1\textwidth]{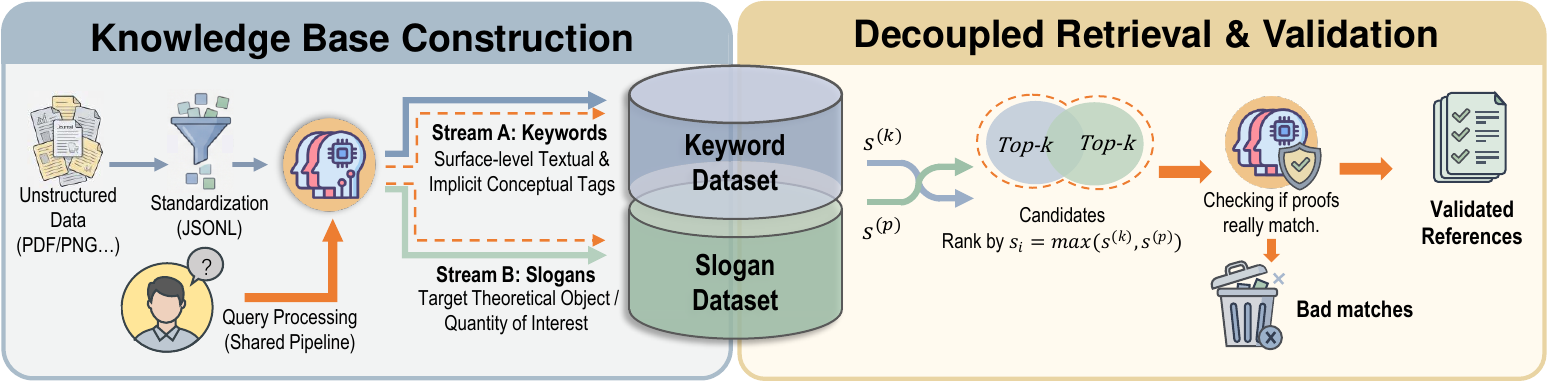}
    \caption{Knowledge construction and dual-channel retrieval for RAG-enhanced statistical reasoning.}
    \label{fig:rag_pipeline}
\end{figure}

In this section, we examine whether the Statistical Research Dataset in \textbf{StatEval} can improve statistical reasoning under a retrieval-augmented generation (RAG) setting, with model parameters kept fixed. We build a tailored knowledge construction and retrieval pipeline, illustrated in Figure~\ref{fig:rag_pipeline}, that converts research-level proof data into structured references and retrieves proof-relevant evidence for downstream multi-step derivations.

\subsection{Dual-Channel Retrieval Mechanism}
\label{subsec:dual_channel_retrieval}

Recent benchmarks show that current retrievers remain limited on reasoning-oriented mathematical retrieval compared with semantic or formula-based tasks \citep{ju2025mirb}. Unlike surface-level retrieval, proof-relevant retrieval must identify transferable proof strategies under precise assumptions and notation. This challenge is particularly salient in statistical proofs, where the key reasoning signal is often distributed across definitions, regularity conditions, and intermediate derivations, and may be further weakened by the positional bias of long-document retrievers \citep{coelho2024dwell,zeng2025empirical}.

To make the retrieval unit more proof-aware, we refine each data instance into a structured representation using an LLM. Specifically, we extract two complementary features: \textit{Keywords}, which capture high-signal mathematical concepts, assumptions, and constraints; and a \textit{Slogan}, a concise academic phrase summarizing the target theoretical object or proof intent. These features provide both constraint-level and strategy-level signals for retrieval.

Retrieving useful references requires balancing formal constraints with proof intent. Since fixed weighted fusion can be unstable when combining heterogeneous signals such as lexical overlap and semantic similarity across queries \citep{bruch2023analysis,hsu2025dat}, we adopt a dual-channel retrieval mechanism. Given a query, we construct keyword embeddings ($Q_k$) and slogan embeddings ($Q_p$), and compare them with candidate embeddings ($D_k, D_p$). After row-wise $\ell_2$ normalization, the base similarity for each channel is computed as $s = \max(\tilde{Q}\,\tilde{D}^{\top})$.
This yields separate scores for the keyword channel ($s^{(k)}_i$) and the slogan channel ($s^{(p)}_i$). To preserve strong evidence from either formal constraint matching or proof-intent alignment, the final ranking score for candidate $i$ is defined as $s_i = \max\left(s^{(k)}_i,\; s^{(p)}_i\right)$.
Thus, a candidate is retained for verification if it exhibits a strong signal in either channel.

Finally, because surface-level similarity does not by itself guarantee mathematical applicability, we introduce a proof-aware verification stage. An LLM evaluator re-ranks the merged top candidates according to reasoning strategy, proof transferability, and domain compatibility rather than semantic similarity alone. This step filters out references that are superficially relevant but mathematically incompatible, ensuring that the retrieved materials provide reliable support for downstream multi-step reasoning.

\subsection{Experiment Setup and Main Results}

\paragraph*{Settings}
Since research-level proofs remain substantially more challenging than foundational statistical tasks, we construct a focused RAG test set of 200 core proof problems sampled from the Statistical Research Dataset branch of StatEval-mini. The baseline performance on this subset is close to the full evaluation results in Table~\ref{tab:research_results}, indicating that the subset preserves the original difficulty and category distribution and can serve as a reliable proxy for studying retrieval augmentation.

We evaluate three representative models: GPT-5.4, Qwen3.6-plus, and Gemini-3.1-Pro. For each, we compare direct generation with a RAG-enhanced setting, using references from our hybrid retrieval module and a proof-oriented structured prompt.

\paragraph*{Prompt Design}
The RAG prompt is designed to encourage proof transfer rather than direct copying. Before using the retrieved references, the model must first analyze the target problem independently. It is then instructed to establish an abstraction barrier and perform isomorphic mapping, extracting transferable algebraic or probabilistic structures from the references and translating them into the notation of the target problem.

To reduce negative transfer, the prompt further requires the model to discard references whose proof strategies conflict with the prerequisite lemmas or hints in the question. The final response includes a reference-usage analysis, a proof blueprint, and the formal proof.

\begin{table}[htbp]
  \centering
  \caption{\textbf{Performance comparison across different models and categories}}
  \begin{threeparttable}
    \begin{tabular}{ll cccccc}
      \toprule
      \multirow{2}[3]{*}{Group} & \multirow{2}[3]{*}{Metric} & \multicolumn{2}{c}{GPT-5.4} & \multicolumn{2}{c}{Qwen3.6-plus} & \multicolumn{2}{c}{Gemini-3.1-Pro} \\
      \cmidrule(lr){3-4} \cmidrule(lr){5-6} \cmidrule(lr){7-8}
      & & Base & RAG & Base & RAG & Base & RAG \\
      \midrule
      \multirow{3}{*}{Difficulty}
      & Easy (n=100) & 77.12 & \textbf{84.59} & 67.44 & \textbf{73.68} & 66.31 & \textbf{72.24} \\
      & Medium (n=50) & 66.03 & \textbf{73.79} & 48.75 & \textbf{61.18} & 52.25 & \textbf{54.35} \\
      & Hard (n=50) & 51.79 & \textbf{61.16} & 33.24 & \textbf{45.98} & 31.18 & \textbf{47.75} \\
      \midrule
      \multirow{8}{*}{Category\tnote{*}}
      & AP (21.0\%) & 56.89 & \textbf{70.31} & 43.50 & \textbf{53.92} & 41.67 & \textbf{50.42} \\
      & GE (18.5\%) & 75.14 & \textbf{78.51} & 59.02 & \textbf{64.85} & 54.83 & \textbf{63.54} \\
      & SG (18.5\%) & 83.85 & \textbf{88.81} & 71.81 & \textbf{77.19} & 72.70 & \textbf{79.09} \\
      & OR (14.0\%) & 60.80 & \textbf{75.49} & 49.17 & \textbf{54.54} & 47.80 & \textbf{56.62} \\
      & CS (11.5\%) & 52.04 & \textbf{57.63} & 36.60 & \textbf{58.44} & 44.45 & \textbf{48.64} \\
      & IC (6.5\%) & 59.52 & \textbf{66.48} & 40.58 & \textbf{54.48} & 44.04 & \textbf{53.42} \\
      & DP (6.0\%) & 89.28 & \textbf{92.46} & 76.54 & \textbf{84.52} & 78.12 & \textbf{81.54} \\
      & TV (4.0\%) & 73.28 & \textbf{81.09} & 63.96 & \textbf{76.56} & 57.87 & \textbf{69.69} \\
      \midrule
      Overall & Mean & 68.01 & \textbf{76.03} & 54.22 & \textbf{63.63} & 54.01 & \textbf{61.65} \\
      \bottomrule
    \end{tabular}
    
    \begin{tablenotes}
      \small
      \item[*] \textit{Category Abbreviations:} \textbf{AP}: Asymptotic Properties; \textbf{CS}: Convergence \& Stability; \textbf{DP}: Distributional Properties; \textbf{GE}: Generalization \& Error Bounds; \textbf{IC}: Identifiability \& Consistency; \textbf{OR}: Optimality Results; \textbf{SG}: Structural Guarantees; \textbf{TV}: Testing Validity. Percentages are computed over the selected 200 questions.
    \end{tablenotes}
  \end{threeparttable}
  \label{tab:rag_main_results}
\end{table}

\paragraph*{Results and Analysis}
As shown in Table~\ref{tab:rag_main_results}, RAG consistently improves all three models across difficulty levels and theoretical categories. The overall mean score increases from 68.01 to 76.03 for GPT-5.4 (+11.8\% relative improvement), from 54.22 to 63.63 for Qwen3.6-plus (+17.4\%), and from 54.01 to 61.65 for Gemini-3.1-Pro (+14.2\%). These gains are especially pronounced on Medium and Hard problems, suggesting that retrieved references help models bridge missing intermediate reasoning steps rather than merely providing surface-level cues.


Across difficulty levels, as shown in Table~\ref{tab:rag_main_results}, RAG consistently improves performance for all three models, with pronounced gains on Medium and Hard problems. On Easy tasks, GPT-5.4 with RAG reaches 84.59, suggesting that, with structured retrieval support, the model can execute relatively complete statistical proof pipelines at a level comparable human experts in mathematical derivations. As difficulty increases, the benefits of retrieval become even more evident: on Medium and Hard settings, RAG boosts the models’ ability to reconstruct missing intermediate arguments and complete multi-step logical dependencies.

Across theoretical categories, the most notable empirical gains appear in tasks that demand dense structural reasoning, with several domains successfully reaching or nearing an elite 90-point threshold. This exceptional performance is largely driven by RAG’s role as an external cache for precise mathematical templates. In categories such as Distributional Properties (DP), where GPT-5.4 with RAG achieves a near-saturated apex of 92.46, statistical proofs often rely on standardized parametric transformations, change-of-variables techniques, or classic moment-generating functions. Since the base model already possesses strong parametric knowledge in this area, the retrieved references act as exact algebraic safeguards, eliminating minor calculation errors and pushing the score past the 90-point mark. 

\begin{figure}[t]
    \centering
    \includegraphics[width=0.92\textwidth,height=0.72\textheight,keepaspectratio]{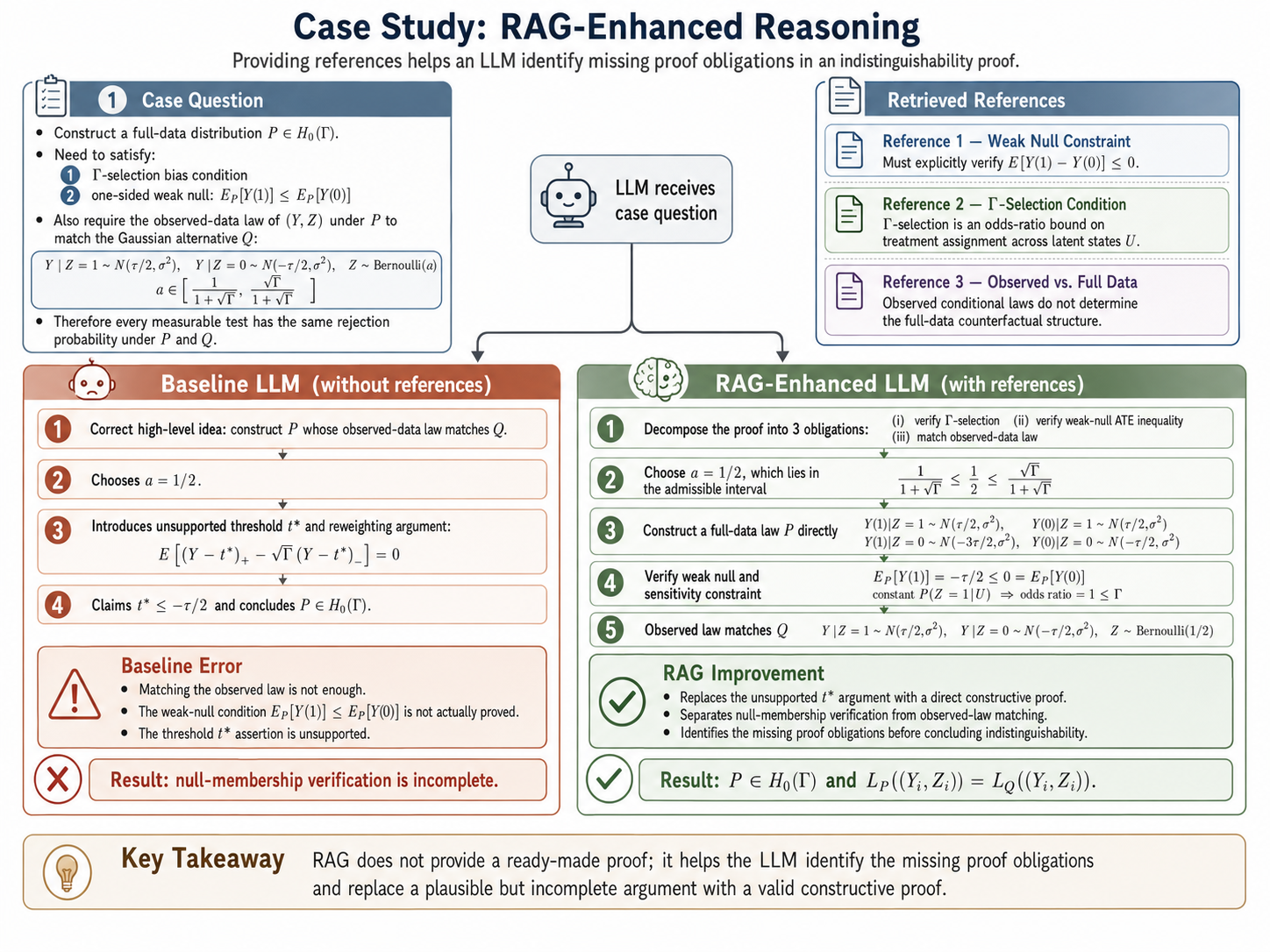}
    \vspace{-0.5em}
    \caption{
    Case study of RAG-enhanced reasoning for an indistinguishability proof under the $\Gamma$-selection sensitivity model\citep{yadlowsky2022bounds}. 
    The baseline matches the observed-data law but leaves null membership incomplete, whereas the RAG-enhanced response identifies and verifies missing proof obligations.
    }
    \label{fig:rag_case_study}
\end{figure}

\paragraph*{Case Study}
Figure~\ref{fig:rag_case_study} presents a representative indistinguishability proof in which retrieval improves the reasoning process of the model. The task is to construct a full-data distribution $P\in H_0(\Gamma)$ whose observed-data law matches that of a Gaussian alternative $Q$. The baseline model captures the high-level idea of matching the observed distribution of $(Y,Z)$, but fails to verify that the constructed distribution belongs to the null class. In particular, it introduces an unsupported threshold argument and does not establish the weak-null condition $\mathbb{E}_P[Y(1)]\le \mathbb{E}_P[Y(0)]$.

With retrieved references, the model decomposes the proof into three necessary obligations: verifying the $\Gamma$-selection constraint, proving the weak-null ATE inequality, and matching the observed-data law. The RAG-enhanced response replaces the unsupported threshold reasoning with a direct constructive proof of a valid full-data law $P$. This example shows that StatEval-Research references improve reasoning by exposing missing proof obligations and organizing the argument around the structural requirements of the hypothesis class.

\section{Application III: Domain-Specific Fine-tuning}
\label{sec:finetuning_app}

In the third application, we investigate whether \textbf{StatEval} can serve as a source of supervision for improving statistical reasoning in LLMs. Due to computational constraints, we adopt \textbf{Qwen3-4B}, a compact open-source model with 4B parameters, as the base model to examine whether a general-purpose LLM can benefit from domain-specific alignment for enhanced reasoning ability using StatEval. Leveraging the difficulty-aware signals in \textbf{StatEval}, we aim to demonstrate that targeted statistical supervision can improve rigorous derivation capabilities even in relatively small models. We adopt a two-stage training paradigm, where the model is first strengthened on general mathematical reasoning tasks and subsequently aligned with specialized statistical derivation data.

\subsection{Phase I: General Mathematical Reasoning}

This stage primarily aims to strengthen the model's general mathematical reasoning before domain-specific statistical alignment. We curate approximately 605k high-quality reasoning samples from: Mathematical Reasoning, including the supervised fine-tuning dataset of NVIDIA Nemotron-4 340B Math v2 \citep{du2025nemotron} and the DASD Thinking Dataset \citep{yan2026distribution}; and Instruction Following, sourced from AM-DeepSeek-R1-Distilled-1.4M \citep{zhao20251}. We prompt gpt-oss-120b to generate medium-length Chain-of-Thought (CoT) solutions and use Qwen2.5-32B-Instruct as an LLM-as-a-judge for verification \citep{moshkov2025aimo}. We further filter responses by context length, structural validity, standardized ``\texttt{<think>}'' formatting, absence of function calls, and repetitive content patterns.

For Supervised Fine-Tuning, we adopt a progressive bucketed training strategy to efficiently handle the varying lengths of reasoning traces. Instead of using a fixed 64K context window from the start, we partition the dataset into length-based buckets and advance the training sequentially in stages (4K--8K--16K $\rightarrow$ 32K $\rightarrow$ 64K). This approach significantly reduces end-to-end training costs without compromising the model's ability to process long-context information at the final target length. Further details on the Phase I dataset composition, response filtering statistics, and training strategy are provided in the supplementary material.

\subsection{Phase II: Domain-Specific Statistical Alignment}

In the second stage, we adapt the model from general mathematical reasoning to domain-specific statistical alignment. This phase targets theoretical proofs and probabilistic derivations using the \textit{Easy} tier of our \textbf{Statistical Research Dataset}. We design a curriculum-driven alignment strategy that combines Rejection Sampling Supervised Fine-Tuning (SFT) and Direct Preference Optimization (DPO), with training data selected according to the model's empirical performance on statistical tasks.

\paragraph*{Difficulty Estimation and Data Stratification}
To construct the training set, we first evaluate the Phase I checkpoint on the target corpus. For each problem, we sample eight candidate responses and compute \textbf{Pass@8}, defined as the proportion of correct solutions among the eight samples. Based on these scores, we stratify the data into three groups:

\begin{itemize}[leftmargin=*]
    \item \textbf{Mastery Filtering (Pass@8 = 1.0)}: Problems solved perfectly are excluded, preventing over-training on mastered patterns and focusing on capability-frontier cases.
    \item \textbf{Synthetic SFT for Hard Samples (Pass@8 $< 0.25$)}: For low-success problems, we use \textbf{GPT-OSS-120B} as a teacher model conditioned on gold-standard answers to synthesize correct reasoning traces for SFT.
    \item \textbf{Preference Alignment for Mid-tier Samples ($0.25 \le \text{Pass@8} < 0.75$)}: For moderately solvable problems, we construct DPO preference pairs \citep{rafailov2023direct} from the model's own correct and incorrect responses, encouraging it to distinguish valid reasoning from common technical errors.
\end{itemize}

\paragraph*{Training Execution}
The alignment process proceeds sequentially. We first perform SFT on synthetic correct responses to inject statistical exposition and proof rigor, and then apply DPO to internal preference pairs to enlarge the margin between correct and incorrect reasoning trajectories. This targeted procedure helps the compact open-weights model reduce logical fallacies and improve the precision of intermediate derivations.

\begin{table}[ht]
\centering
\caption{Performance of domain-specific statistical alignment on the statistical research dataset.}
\label{tab:statistical_alignment}
\begin{tabular}{lccc} 
\toprule
\textbf{Model} & \textbf{Easy} & \textbf{Medium} & \textbf{Hard} \\ \midrule
Qwen3-4B & 28.55 & 23.96 & 17.37 \\
Qwen3-4B + Phase I & 30.12 & 21.47 & 18.14 \\
Qwen3-4B + Phase I + Phase II & \textbf{36.28} & \textbf{27.32} & \textbf{22.91} \\ \bottomrule
\end{tabular}
\vspace{-1em}
\end{table}

\paragraph*{Results and Analysis}

The quantitative results validate our domain-specific statistical alignment framework. As shown in Table~\ref{tab:statistical_alignment}, the introduction of Phase I yields marginal improvements on \textit{Easy} and \textit{Hard} tasks, yet drops noticeably on the \textit{Medium} split, with the score degrading from 23.96 to 21.47. This empirical pivot underscores that general mathematical fluency alone is insufficient and does not inherently transfer to rigorous statistical derivation; instead, naive general tuning may even induce distributional misalignment on moderately complex tasks.

In contrast, integrating Phase II, which is explicitly guided by the difficulty-aware curriculum structure of StatEval, unlocks substantial and consistent gains across all proficiency levels. Compared to the vanilla Qwen3-4B baseline, the complete alignment pipeline yields a 27.08\% relative improvement on Easy tasks, a 14.02\% relative improvement on Medium tasks, and a 31.89\% relative improvement on Hard tasks. These compelling margins demonstrate that StatEval can provide actionable, highly targeted supervision for domain alignment and acts as an effective diagnostic and training resource for fortifying statistical reasoning.


Although our experiments are conducted on the compact Qwen3-4B model due to computational constraints, the observed gains suggest that the same domain-specific alignment pipeline may also benefit larger foundation models. We leave systematic investigation of this scaling behavior to future work with more substantial computational resources.

\section{Conclusion}
\label{sec:conclusion}

While current LLM evaluations largely overlook statistical reasoning, we introduce \textbf{StatEval} to bridge this gap. This comprehensive benchmark comprises over 106,000 undergraduate and research-level problems, accompanied by a standardized pipeline for extracting and constructing statistical datasets from unstructured text. Our evaluations reveal that although state-of-the-art models struggle with research-level tasks, integrating RAG and targeted fine-tuning significantly improves their statistical reasoning. Ultimately, StatEval establishes a rigorous foundation for advancing research-oriented statistical AI tools.



\section{Data Availability Statement}\label{data-availability-statement}

The full datasets are not publicly available yet. However, a sample dataset is provided on Hugging Face (\url{https://huggingface.co/datasets/StatAILab/StatEval}) to verify our results.

\bibliography{bibliography.bib}

\end{document}